\documentclass[letterpaper, 10 pt, conference]{ieeeconf}  

\IEEEoverridecommandlockouts                              

\usepackage{url}
\usepackage{amsmath}
\usepackage{algorithm}
\usepackage{algorithmic}
\usepackage[colorinlistoftodos,prependcaption,textsize=small]{todonotes} 
\usepackage{array,booktabs,threeparttable}
\usepackage{subcaption}
\usepackage{bbold}
\usepackage{adjustbox}
\usepackage[utf8]{inputenc}
\usepackage{tikz}
\usepackage{pgfplots}
\usepackage{mathtools}
\usepackage{soul}
\usepackage{lipsum}
\usepackage{multirow}

\pgfplotsset{compat=newest}

\makeatletter
\let\NAT@parse\undefined
\makeatother
\usepackage{hyperref}
\hypersetup{ 
  colorlinks,
  citecolor=green,
  linkcolor=red,
  urlcolor=blue
}
\usepackage[capitalise,noabbrev]{cleveref}  

\graphicspath{{figures/}}

\newlength\figureheight  
\newlength\figurewidth  

\usepackage[figurename=Fig.]{caption}

\title{\LARGE \bf
epBRM: Improving a Quality of 3D Object Detection using End Point Box Regression Module  
}

\author{Kiwoo Shin$^\dagger$, and Masayoshi Tomizuka$^\dagger$
\thanks{The authors are with the department of Mechanical Engineering, University of California, Berkeley, CA 94720, US. {\tt\footnotesize \{kiwoo.shin, tomizuka\}@berkeley.edu}}
\thanks{$^\dagger$Mechanical Systems Control Lab, University of California, Berkeley, CA, USA.}%
\thanks{This work was in part supported by Berkeley Deep Drive. Kiwoo Shin is supported by Samsung Scholarship.}
}

\begin{document}

\maketitle
\thispagestyle{empty}
\pagestyle{empty}

\begin{abstract}
We present an endpoint box regression module(epBRM), which is designed for predicting precise 3D bounding boxes using raw LiDAR 3D point clouds. The proposed epBRM is built with sequence of small networks and is computationally lightweight. Our approach can improve a 3D object detection performance by predicting more precise 3D bounding box coordinates. The proposed approach requires 40 minutes of training to improve the detection performance. Moreover, epBRM imposes less than 12ms to network inference time for up-to 20 objects.

The proposed approach utilizes a spatial transformation mechanism to simplify the box regression task. Adopting spatial transformation mechanism into epBRM makes it possible to improve the quality of detection with a small sized network.

We conduct in-depth analysis of the effect of various spatial transformation mechanisms applied on raw LiDAR 3D point clouds. We also evaluate the proposed epBRM by applying it to several state-of-the-art 3D object detection systems.

We evaluate our approach on KITTI dataset\cite{geiger_vision_2013}, a standard 3D object detection benchmark for autonomous vehicles. The proposed epBRM enhances the overlaps between ground truth bounding boxes and detected bounding boxes, and improves 3D object detection. Our proposed method evaluated in KITTI test server outperforms current state-of-the-art approaches.

\end{abstract}

\section{INTRODUCTION}

3D object detection systems become a core component in most of recent autonomous vehicles system. Autonomous vehicles are required to precisely detect and track the surrounding objects in real-time to achieve safe driving. Among the various sensors used in recent autonomous vehicles system, LiDAR which measures distance between objects and the ego-vehicle become a crucial sensor for 3D object detection system.  

\begin{figure}[t]
    \centering
    \includegraphics[width=0.75\linewidth]{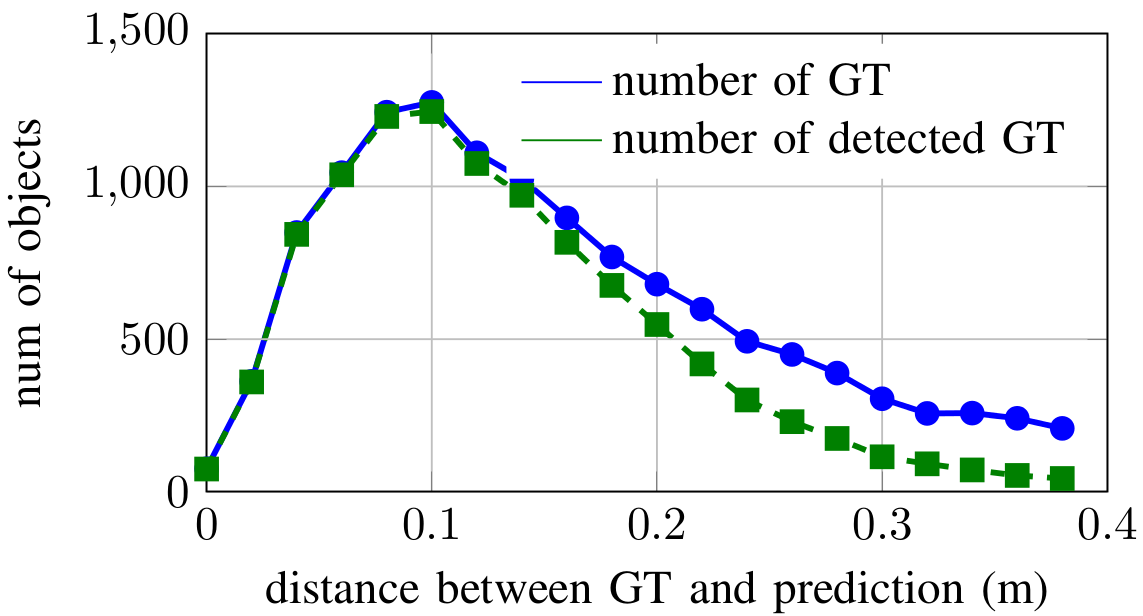} 
    \caption{Many of predictions located close to ground truth objects(: GT) have 3D overlap lower than 70\%. 'Detected' means that overlap between GT and prediction is greater than 70\%.}
    \label{fig:iou_distribution}
\end{figure}

The main goal of this work is to improve quality of bounding box predictions in order to increase recall of 3D object detection from LiDAR 3D point clouds. Despite the close proximity between predicted and ground truth objects, recent 3D object detection systems perform below a specific requirements, thereby increasing false-positive detections.(see Fig. \ref{fig:iou_distribution})

\begin{figure}[t]
    \centering
    \includegraphics[width=0.8\linewidth]{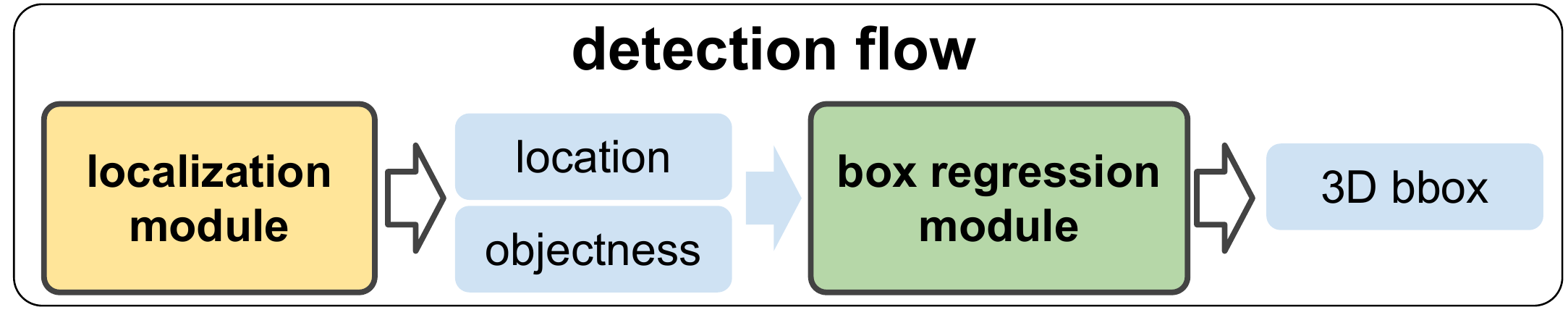} 
    \caption{Decomposing a detection pipeline into localization module and box regression module allows us a flexibility in building detection system as well as enhanced efficiency in training box regression task.}
    \label{fig:big_picture}
\end{figure}

\begin{figure*}[t]
  \centering
  \includegraphics[width=0.8\linewidth]{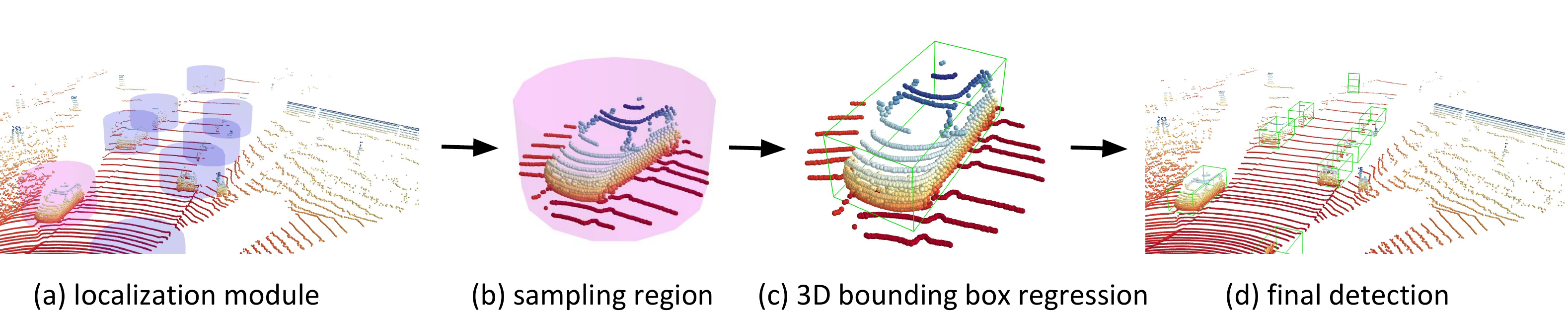}
  \caption{3D bounding box regression pipeline. (a) The localization module predicts the location of objects. (b) samples point clouds at each predicted location, (c) and predicts 3D bounding box which fits to the object. (d) shows final detection results.}
  \label{fig:regression_pipeline}
\end{figure*}

In autonomous driving, the size of search space is usually 80 m(side) X 70 m(forward) from ego vehicle, which is very huge compared to the size of objects, such as car, pedestrian and cyclist. Moreover, the number of object, which can be used as training samples, is also limited. Thus, the majority of search space is occupied by non-object area. During training process, the network is mostly trained to classify object/non-object areas(localization task) and less effort is made for predicting precise 3D bounding boxes(box regression task). Previous works attempt to overcome such challenge by using efficient encoding of 3D bounding box\cite{ku_joint_2017} and/or by adopting focal loss to dynamically control weight of each task\cite{Tsung_focal_2017}, but those approaches have not been very successful.


The main idea behind our proposed detection system is to decompose the whole detection pipeline into a localization task and a box regression task and train each task independently (Fig. \ref{fig:big_picture}). By decomposing the detection pipeline into two tasks, a box regression network can be trained more efficiently, 1) with a network specifically designed for regression task(\cref{sec:spatial}), and 2) using training samples containing information about 3D bounding boxes only.(\cref{sec:sampling}).


Because individual network transition slows the detection speed, most previous 3D object detection systems are developed from a unified structure(\cite{ku_joint_2017, yang_pixor_2018, lang_pointpillars_2018, li2016vehicle}). Therefore, in our approach, box regression network is implemented to accelerate real time object detection. To meet such requirement, we design the epBRM with a small size network, which imposes only 12ms overhead for up to 20 objects(\cref{sec:loss}).



We use current 3D object detection systems as localization module in our detection pipeline. In this work, a localization module is responsible for predicting the location of object and its corresponding confidence score. And epBRM is responsible for predicting all coordinates required for a 3D bounding box, including location, rotation and size of object (Fig. \ref{fig:regression_pipeline}). 

We evaluate the proposed method by applying to the result of current 3D object detection systems, AVOD(FPN)\cite{ku_joint_2017}, F-PointNet\cite{qi_frustum_2018} and PointPillars\cite{lang_pointpillars_2018}. The evaluation result demonstrates that the proposed approach improves the 3D object detection performance immediately after 10k of training iteration, which takes only 40 minutes using one Titan X GPU(not pascal) and i7-6700k CPU.

\section{Related works}


\subsection{Bird's eye view based approaches} \label{sec:rel_bev}
In early works in 3D object detection system, input feature is represented by projecting LiDAR 3D point clouds onto 2D planes and used CNN-based detection network for prediction\cite{luo_fast_2018,simon_complex-yolo_2018,yang_pixor_2018}. A core problem shared by these approaches is a loss of information by projection. Thus, these methods showed low performance in detecting small objects such as pedestrian and cyclist. 

\subsection{Sensor fusion based approaches} \label{sec:rel_fusion}
The density of LiDAR 3D point clouds is sparse when the object is far from the LiDAR. Several sensor fusion based approaches have been proposed to utilize 2D RGB images that can provide higher resolution than LiDAR\cite{cvpr17chen, ku_joint_2017, liang_deep_2018,enzweiler2011multilevel,gonzalez2017board,zining2018fusing}. These methods concatenate features extracted from 2D images and LiDAR point clouds to generate input feature for the region proposal network. These approaches improved detection performance for small objects, but they required additional task for synchronization and calibration between multi-sensors.

\subsection{3D based approaches} \label{sec:rel_3D}
One approach in 3D object detection system is to convert Lidar 3D point clouds into voxels and extracts voxelwise features for predicting 3D bounding boxes\cite{zhou2017voxelnet, shi2018pointrcnn, li20173d, graham2014spatially, lang_pointpillars_2018, vora2019pointpainting, yoo20203d}. Several works proposed feature representation based on 3D CNN, but they required high computation\cite{xiaodeep, li2016vehicle}.

Another line of work in 3D object detection systems can be categorized between sensor fusion based approach and 3D based approach, which is to use 2D object detection results as region proposals and apply PointNet\cite{qi_pointnet_2017, qi_pointnet++_2017} to predict 3D bounding boxes\cite{qi_frustum_2018, shin2019roarnet}. One drawback behind these approaches is a slow inference speed due to the need for two sequentially connected pipelines, one for 2D image detection and the other one for 3D point clouds detection.

\section{3D Box Regression Module}


\subsection{Role and goal of epBRM}

The main purpose of epBRM is not to discover objects that are missed by localization module, but to predict precise 3D bounding box coordinates by using point clouds sampled around the location predicted from the localization module. Therefore, we set a target distance($dist_{\text{bound}}$), which represents the maximum distance between a ground truth object and the localization module's predicted location of the object, further refining a 3D bounding box. For objects located more than $dist_{\text{bound}}$ from the ground truth object, we do not aim for such detections to be further regressed.

There are two main factors for deciding the value of $dist_{\text{bound}}$: 1) performance of the localization module, and 2) the representation ability of epBRM.

First, $dist_{\text{bound}}$ is inversely related to the precision of the localization module. For example, if the localization module is capable of predicting precise location of objects, then epBRM only needs to work well for samples with small error. In this case, we assign small value for $dist_{\text{bound}}$.

Second, the value of $dist_{\text{bound}}$ is also related to the representation ability of a network used for epBRM. If we increase the value of  $dist_{\text{bound}}$, it also increases the complexity of task that epBRM is responsible for. If we constrain the representation ability of epBRM for computational efficiency, the value of $dist_{\text{bound}}$ will also decrease.

In \cref{sec:exp2}, we evaluate the effect of  $dist_{\text{bound}}$ on various localization modules and its effect on detection performance.

\subsection{epBRM with spatial transformation mechanism} \label{sec:spatial}

Aiming for better computational efficiency and real time inference, we restrict the size of the network for epBRM at the cost of representation ability of the network. This necessitates a need for an additional method to simplify the box regression task in order for a network with low representation ability to be capable of predicting precise 3D bounding boxes. 



To avoid voxelizing\cite{zhou2017voxelnet} and/or grouping 3D point clouds\cite{qi_pointnet++_2017} which require extra processing time, we adopt spatial transformation mechanism\cite{jaderberg2015spatial} into epBRM to get a similar effect with pooling based approaches. In essential, our approach first spatially transforms input point clouds and then predicts 3D bounding boxes. We then take the inverse transform from the predicted 3D bounding box to the input point cloud's original coordinate.



In this work, we are focusing on four spatial transformation mechanisms: translation, rotation, scaling, and centering. They take the input point clouds $P \in \mathbb{R}^{\text{Nx3}}$ with N point clouds, and 3 dimensions for x, y, z coordinates.

\textbf{Translation and Centering} The output of transformation mechanism is defined by ($t_{x}, t_{y}, t_{z})$ which represents the translation of $P$ by $x_{0}, y_{0}, z_{0}$ as following equations. 

\begin{align}
\label{eqn:t_translation}
\begin{split}
 x_{0} &= \text{2}* ( \sigma(t_{x}) \text{ - 0.5}) * T_{x} ,
\\
 y_{0} &= \text{2}* ( \sigma(t_{y}) \text{ - 0.5}) * T_{y} ,
\\
 z_{0} &= \text{2}* ( \sigma(t_{z}) \text{ - 0.5}) * T_{z}
\end{split}
\end{align}
where $\sigma(\cdot)$ is the sigmoid activation function. In this work, we set $(T_{x}, T_{y}, T_{z})$ to the same values as $dist_{\text{bound}}$. 

The difference between translation mechanism and centering mechanism is determined by whether there exists target transformation parameters to be learned. In translation mechanisms, the network is trained without target transformation parameters to extract most informative feature(\cref{eqn:loss1}). Unlike to the translation mechanism, centering mechanism is trained to predict the center of object as transformation parameters(\cref{eqn:loss2}).

\textbf{Rotation} The output of rotation mechanism is defined by $t_{r}$ which represents a clockwise rotation of $P$ around z axis by $\text{rot}_{z}$ as following equation. 

\begin{align}
\label{eqn:t_rotation}
\begin{split}
 \text{rot}_{z} &= \text{2}* ( \sigma(t_{r}) \text{ - 0.5}) * T_{r}
\\
\end{split}
\end{align}
We set the value of $T_{r}$ as $\pi$/4 to avoid excessive rotation of input point clouds.

\textbf{Scaling} The output of scaling mechanism is defined by $(t_{s, xy}, t_{s, z})$ which represent the scaling of $P$ on ground plane, $\text{scale}_{xy}$ and on z axis, $\text{scale}_{z}$ as following equation.
\begin{align}
\label{eqn:t_scale}
\begin{split}
 \text{scale}_{xy} &= \text{pow}(T_{s, xy}, 2*( \sigma(t_{s, xy}) \text{ - 0.5})),
\\
 \text{scale}_{z} &= \text{pow}(T_{s, z}, 2*( \sigma(t_{s, z}) \text{ - 0.5}))
\end{split}
\end{align}
We set a value of $T_{s, xy}, T_{s, z}$ as 2.0, thus constrain scaling factors within [$\frac{1}{2}$, 2] to avoid distorting input point clouds.







\subsection{3D bounding box regression} \label{sec:block}

After the transformation is applied to input point clouds, epBRM finally predicts coordinates for the 3D bounding boxes, including location, rotation, and size of objects. We use confidence score predicted from each localization module for scoring our predicted 3D bounding boxes.

\textbf{Location} is predicted by 3 coordinates $(t_{x}, t_{y}, t_{z})$ for $x, y, z$ direction relative to the origin as following equation:

\begin{align}
\label{eqn:eq_localization}
\begin{split}
 x &= \text{2}* ( \sigma(t_{x}) \text{ - 0.5}) * d_{x} ,
\\
 y &= \text{2}* ( \sigma(t_{y}) \text{ - 0.5}) * d_{y} ,
\\
 z &= \text{2}* ( \sigma(t_{z}) \text{ - 0.5}) * d_{z}
\end{split}
\end{align}
This has a same form as Equation (1) except for the value of the hyper parameters, $(d_{x}, d_{y}, d_{z})$.
Aiming for finer prediction of location, we set values for the $(d_{x}, d_{y}, d_{z})$ as follows:
\begin{align}
\label{eqn:refine}
\begin{split}
 [d_{x}, d_{y}, d_{z}]^{\text{T}} & = \text{0.5}* [T_{x}, T_{y}, T_{z}]^{\text{T}}
\\
\end{split}
\end{align}

\textbf{Rotation} is predicted based on hybrid formulation of classification and regression methods. We equally divide [0', 180'] to $N_{R}$ bins and the rotation angle is predicted by 2x$N_{R}$ coordinates ($t_{r\_\text{cls}(i)}, t_{r\_\text{reg}(i)})_{i=1}^{N_{R}}$.

\textbf{Size} is predicted by 3 coordinates, $(t_{h}, t_{w}, t_{l})$, for height($h$), width($w$), and length($l$). We set anchor of size coordinates, $(h_{a}, w_{a}, l_{a})$. We use (1.50m, 1.57m, 3.33m) for car, (1.73m, 0.6m, 0.8m) for pedestrian and (1.73m, 0.6m, 1.76m) for cyclist. Then, the size of object is predicted by:

\begin{align}
\begin{split}
h = h_{a} e^{t_{h}}, \quad w = w_{a} e^{t_{w}}, \quad l = l_{a} e^{t_{l}}
\end{split}
\end{align}

\subsection{Network structure}

\begin{figure}
    \centering
    \includegraphics[width=0.75\linewidth]{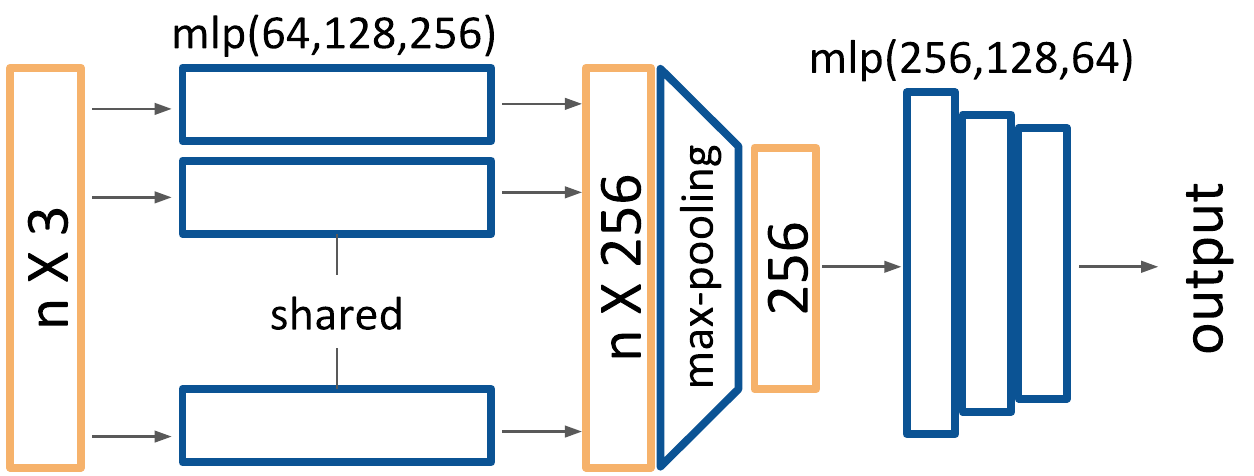} 
    \caption{the modular network structure used as a building block of epBRM. The same structure is used for both transformation mechanism and the 3D bounding box regression task.}
    \label{fig:basic_network}
\end{figure}

We design epBRM as sequence of multiple PointNet building blocks. Fig. \ref{fig:basic_network} shows a structure of the building block based on PointNet . This structure is shared by transformation mechanism and 3D bounding box regression task except for the number of output at the last layer. 

In the case of the transformation mechanism, we transform point clouds using output of network, which is a transformation parameter to be applied to input point clouds(\cref{sec:spatial}) and sample point clouds inside sampling region(\cref{sec:sampler}).

\subsection{Generating training samples} \label{sec:sampling}
We aim for setting the sampling region as tightly as possible around the object in order to exclude redundant point clouds that do not belong to the object. As we only have location prediction from localization module, we need rotation-invariant sampling region. To satisfy both requirements, we use cylinder shaped sampling region in this work.



Considering a prediction error from localization module, we set size of sampling region slightly larger than ordinary size of objects. We sample point clouds inside a cylinder defined by radius $r$ and min/max value of point clouds ($z_{min}, z_{max}$) along z axis. (\cref{tab:samplingregion})

\begin{table}[h]
\caption{the size of sampling region for each class}
\begin{center}
\begin{tabular}{@{}>{\centering}m{1.7cm}|cm{1.7cm}m{1.7cm}@{}}
\hline
class & radius(r) & min height($z_{min}$) & max height($z_{max}$)\\
\hline
Car & 2.4 & -0.5 & 2.5\\
Pedestrian & 0.35 & -0.5 & 2.5\\
Cyclist & 0.8 & -0.5 & 2.5\\
\hline
\end{tabular}
\end{center}
\label{tab:samplingregion}
\end{table}

We then describe data augmentation methods for generating training samples. Assume that the 3D bounding box coordinates of the ground truth object are given by ${loc}_{\text{gt}} = ({loc}_{\text{gt,x}}, {loc}_{\text{gt,y}}, {loc}_{\text{gt,z}})$, ${size}_{\text{gt}} = ({h}_{\text{gt}}, {w}_{\text{gt}}, {l}_{\text{gt}}) $ and ${rot}_{\text{gt}}$.

First, we sample all point clouds inside sampling region centered at $\text{loc}_{\text{gt}}$. Then, we translate sampled point clouds by subtracting $\text{loc}_{\text{gt}}$ from each point and rotate them by ${\text{-}rot}_{\text{gt}}$ to align the object along y axis. 


We change the size of object by multiplying $s_{x}, s_{y}, s_{z}$ for each $x, y, z$ axis of point clouds independently. $s_{x}, s_{y}, s_{z}$ are independently sampled from uniform distribution on the interval [0.9, 1.1]. 



We then rotate the point clouds back to original heading angle plus small angle $r_{z}$ randomly sampled from uniform distribution on the interval $[-pi/8, pi/8]$. 



Finally, we sample ${loc}_{\text{x}}, {loc}_{\text{y}}, {loc}_{\text{z}}$ independently from uniform distribution on the interval $[\text{-}dist_{\text{bound}}, dist_{\text{bound}}]$.



\subsection{Loss, training and runtime} \label{sec:loss}
During the training of each network, we optimize the following multi-task loss end-to-end:
\begin{align}
 Loss &=  L_{\text{loc}} + L_{\text{rot-cls}} + \mathbb{1}^{\text{rot-cls}}[L_{\text{rot-reg}}] +
L_{\text{size}} \label{eqn:loss1}\\
&+L_{\text{loc\_center}} \label{eqn:loss2}
\end{align}
If we use transformation mechanism such as translation, rotation and scaling, then we use \cref{eqn:loss1} as a loss function. If we adopt centering mechanism in epBRM, then we use \cref{eqn:loss2} as a loss function. 


$L_{\text{loc\_center}}, L_{\text{loc}}, L_{\text{rot-reg}}$, and $L_{\text{size}}$ are regression loss for intermediate center location prediction, location, rotation and size of bounding box regression, which are represented as Huber loss. $L_{\text{rot-cls}}$ is classification loss for rotation, which is represented as cross-entropy loss.

For training a network, we use the generated samples described in \cref{sec:sampling}. We use batch of 512 samples with fixed learning rate of 5e-4. We do not apply non-maximum suppression on the predictions.

We use one i7-6700k CPU and one Titan X GPU for training and inference. Up to 20 objects, it takes 6.5ms for sampling point clouds at each location predicted from localization module. For actual network inference, it takes 5.5ms with one transformation mechanism adopted in epBRM.







\section{Experiments}

\subsection{Experiment setup} \label{sec:exp_setup}

\textbf{Dataset} We use KITTI dataset, the 3D object detection benchmark, to evaluate our approach. It provides synchronized 2D RGB images and 3D LiDAR point clouds, carefully calibrated with annotations on car, pedestrian, and cyclist class. We mainly evaluate our method on the car class which has the most training samples. For pedestrian and cyclist, we discuss in \cref{sec:exp3}.


To evaluate our method, we split whole training set into {\tt train} set of 3,717 frames and {\tt val} set of 3,769 frames frames. Frames in {\tt train} set and frames in {\tt val} set are extracted from different video clips. 3D object detection performance is evaluated at 0.7 IoU threshold for car and 0.5 IoU threshold for pedestrian and cyclist.

\textbf{Localization module} We utilize previous 3D object detection systems as localization modules. To show the applicability of the proposed method, we select three different 3D object detection systems publicly available as open source: F-PointNet, AVOD(FPN), and PointPillars. These 3D object detection systems use different feature and method for localization.(\cref{tab:3Ddetection}). We follow official instruction provided by the authors to train the localization modules and freeze it when we train/evaluate the epBRM network.

\begin{table}[t]
\caption{Comparison of 3D object detection systems. Explanations on AVOD(FPN) and F-PointNet are from \cite{feng2019deep}}
\begin{center}
\begin{tabular}{@{}>{\centering}m{1.4cm}|m{2.0cm}m{2.0cm}m{2.0cm}@{}}
\toprule
method &  AVOD(FPN)\cite{ku_joint_2017} & F-PointNet\cite{qi_frustum_2018} & PointPillars\cite{lang_pointpillars_2018}\\
\hline
sensors & LiDAR, camera & LiDAR, camera & LiDAR only\\
\hline
method for localization & fused LiDAR and image feature & pretrained image detector & LiDAR feature\\
\bottomrule
\end{tabular}
\end{center}
\label{tab:3Ddetection}
\end{table}





\textbf{Evaluation metric} To evaluate our approach, we mainly focus on whether epBRM improves a quality of 3D bounding box prediction. As well as recall and mean average precision(mAP) of detection which are general evaluation metrics for 3D object detection benchmark, we also measure \textbf{ratio} of detected ground truth objects defined as follows:
\begin{align}
\label{eqn:ratio}
\begin{split}
 \text{ratio} & :=  \frac{\text{num. of detected GT objects}}{\text{num. of all GT objects}}
\end{split}
\end{align}
where the term 'GT' represents ground truth and 'detected' means that the overlap between ground truth 3D bounding box and predicted 3D bounding box is greater than 70\%.

This measurement considers all ground truth object regardless of its difficulty level and visible size of object in RGB image plane, thus provides more reliable measurement.

\subsection{Comparisons of transformation mechanism} \label{sec:exp1}

We first evaluate each spatial transformation mechanism on 3D bounding box regression task. In this experiment, we utilize PointPillars\cite{lang_pointpillars_2018} as a localization module. We train epBRM with each transformation mechanism for 20k iteration. We use the box regression module which is trained without spatial transformation mechanism as baseline experiment. 

\begin{table}[t]
\caption{a comparison of transformation mechanisms. The recall and mAP are evaluated in moderate level($\dagger$). The ratio is evaluated for all difficulty levels($\ddagger)$ following \cref{eqn:ratio}.}
\begin{center}
\begin{tabular}{@{}>{\centering}m{3.5cm}|cccc@{}}
\hline
transformation mechanism & recall($\dagger$) & mAP($\dagger$) & ratio($\ddagger$)\\
\hline
none & 85.0 & 75.52 & 73.23 \\
translation  & 85.0 & 76.65 & 75.30\\
center  & 85.0 & 77.28 & 76.62\\
translation+rotation  & 82.5 & 73.73&72.97\\
center+rotation  & 85.0 & 75.90 & 75.45\\
center+scale& 85.0 & 76.95 & 75.34 \\
\hline
PointPillars\cite{lang_pointpillars_2018} & 82.5 & 76.29 & 73.17 \\
\hline
\end{tabular}
\end{center}
\label{tab:exp1}
\end{table}

\cref{tab:exp1} indicates that spatial transformation mechanisms is essential component of the box regression module to improve the detection performance. Among the various spatial transformation mechanisms, the centering mechanism is the most effective in improving the detection quality.

\begin{figure}[t]
    \centering
    \includegraphics[width=0.75\linewidth]{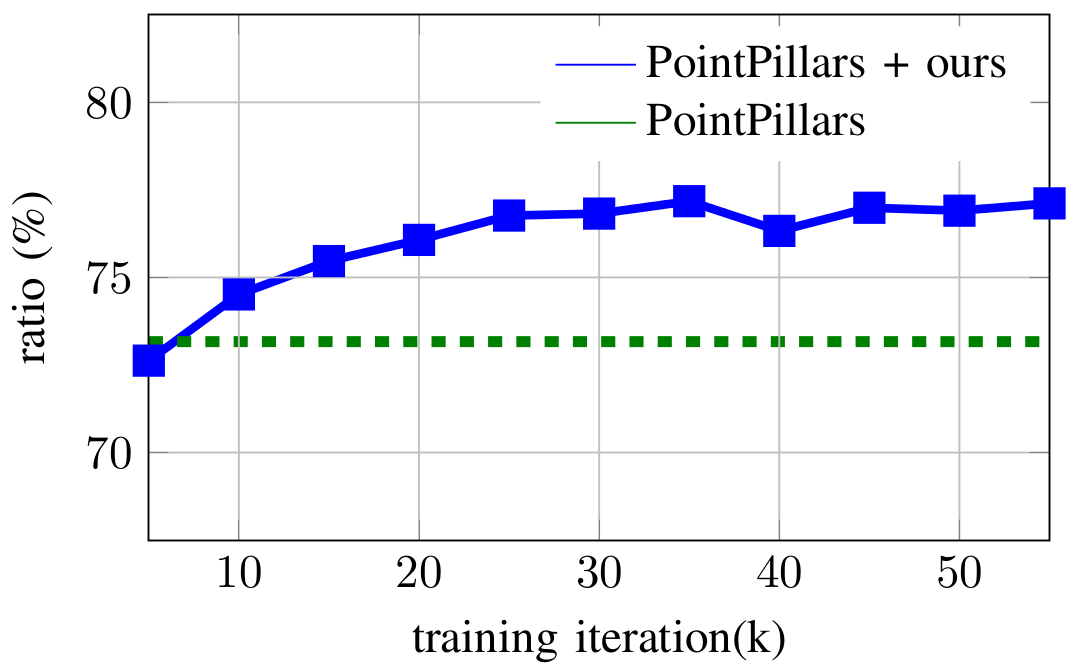} 
    \caption{Training progress of epBRM. Evaluation of car class in KITTI $\tt{val}$ set.}
    \label{fig:training_progress}
\end{figure}

Fig. \ref{fig:training_progress} shows the validation result of centering mechanism until 55k training iterations. Note that epBRM outperforms the PointPillars\cite{lang_pointpillars_2018} after 10k of training iteration. It takes approximately 40 minutes for each 10k iteration of training.

\subsection{Effect of dist-bound} \label{sec:exp2}

In this experiment, we compare effect of $dist_{\text{bound}}$ on several localization modules. The purpose of this experiment is to find optimal value of $dist_{\text{bound}}$ that maximizes the performance of epBRM when epBRM is applied to different localization modules. We again train epBRM with centering mechanism for 20k iteration.

\begin{table*}[t]
\caption{Comparisons of 3D object detection performance on car, pedestrian and cyclist class before/after applying epBRM to each localization module. The performances are evaluated at KITTI \textit{val} set. The recall and mAP are evaluated in moderate difficulty level and the ratio is evaluate in all difficulty level.}
\begin{center}
\begin{tabular}{@{}>{\centering}m{1.7cm}|m{0.8cm}m{0.8cm}m{0.8cm}m{0.8cm}|m{0.8cm}m{0.8cm}m{0.8cm}m{0.8cm}|m{0.8cm}m{0.8cm}m{0.8cm}m{0.8cm}@{}}
\hline
 & \multicolumn{4}{c}{car} &  \multicolumn{4}{c}{pedestrian} &  \multicolumn{4}{c}{cyclist}\\
 \hline
 methods & dist (m) & recall (\%) & mAP (\%) & ratio (\%) & dist (m) & recall (\%) & mAP (\%) & ratio (\%) & dist (m) & recall (\%) & mAP (\%) & ratio (\%) \\
 \hline
 AVOD(FPN) & - &  77.5	& 68.50	& 63.18 & - &  52.5	& 39.78	& 41.36 & - &  55.0	& 37.14	& 45.80 \\
 + ours & 0.15 &  82.5	& 77.02	& 68.24 & 0.15 &  55.0	& 41.54	& 44.04 & 0.15 		&  57.0	& 37.61	& 48.38 \\
 \hline
 F-PointNet & - &  82.5	& 71.36	& 62.06 & - &  72.5	& 55.40	& 59.12 & - &  75.0	& 53.85	& 62.91 \\
 + ours & 0.30 &  82.5	& 72.90	& 65.99 & 0.15 &  77.5	& 63.06	& 66.84 & 0.15 		&  77.5	& 59.92	& 68.09 \\
 \hline
 PointPillars & - &  82.5	& 76.29	& 73.17 & - &  77.5	& 58.95	& 69.30 & - & 77.5	& 61.22	& 62.60 \\
+ ours & 0.15 &  85.0	& 77.28	& 76.62 & 0.15 & 82.5 & 61.65	& 73.29 & 0.15 		&  77.5	& 61.07	& 65.00 \\
\hline
\hline
\end{tabular}
\end{center}

\label{tab:exp2}
\end{table*}
 
\begin{figure}[t]
    \centering
    \includegraphics[width=0.75\linewidth]{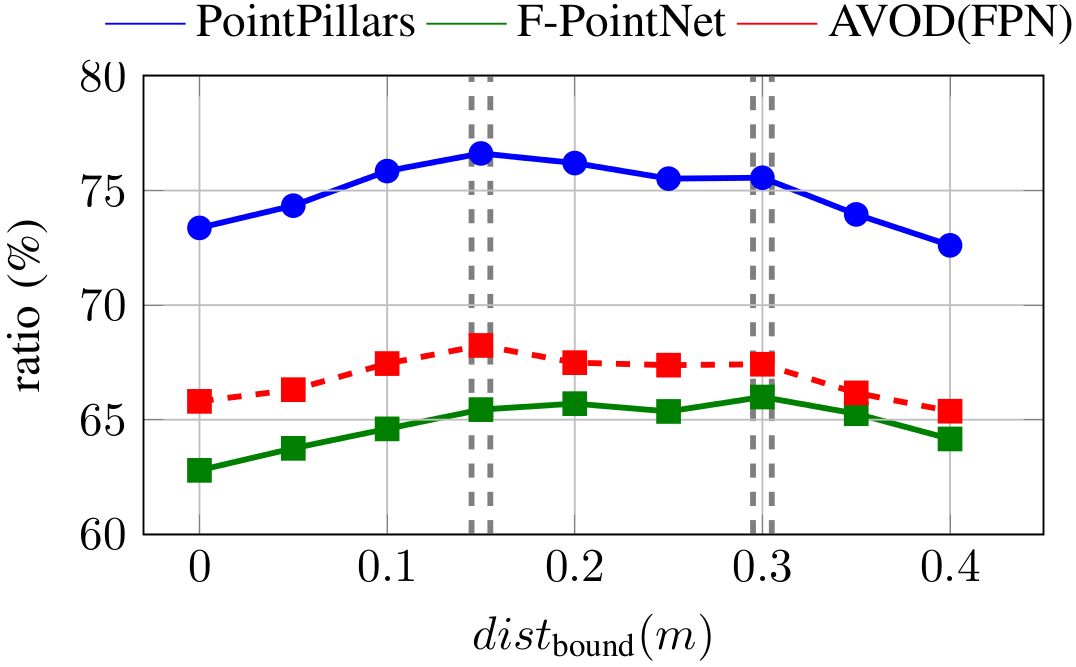} 
    \caption{Effect of $dist_{\text{bound}}$ on each localization module.}
    \label{fig:dist_bound}
\end{figure}

Fig. \ref{fig:dist_bound} visualizes the evaluation results when epBRM is applied to each localization module at different $dist_{\text{bound}}$. For PointPillars and AVOD(FPN), the value of ratio(\cref{eqn:ratio}) peaks when $dist_{\text{bound}}$ is 0.15 and then gradually decreases after that. For F-PointNet, the value of ratio peaks when $dist_{\text{bound}}$ is 0.30.

\cref{tab:exp2} reports the performance gains when epBRM is applied to each localization module at optimal $dist_{\text{bound}}$ found by this experiment.

Note that the performance gain from epBRM significantly differs by each localization module. For example, epBRM greatly improves the performance of AVOD(FPN) while less improvement is observed from F-PointNet. This result infers that the box regression of AVOD(FPN) is comparably worse than the F-PointNet.


 
 

\subsection{Smaller objects: pedestrian and cyclist} \label{sec:exp3}

Our approach is also applicable for predicting precise 3D bounding boxes for smaller objects such as pedestrian and cyclist. Again, we train epBRM using fixed value of $dist_{\text{bound}}$ as 0.15 for 20k iteration and evaluate the performance by applying it to each localization module. Here, we omit the process of finding optimal value of $dist_{\text{bound}}$.

\subsection{Evaluation using KITTI test set} \label{sec:exp4}

We also evaluate the proposed approach by using KITTI \textit{test} set by submitting detection result to KITTI test server. Among several localization modules we evaluate so far, PointPillars\cite{lang_pointpillars_2018} shows the best performance in all categories, thus we apply epBRM to prediction result from PointPillars. We train epBRM for 40k iterations, which takes approximately 160 minutes.

\begin{table}[t]
\caption{Evaluation results for car, pedestrian and cyclist at KITTI \textit{test} set. ($\dagger$) represents reproduced result from our side. ($\ddagger$) represents performance of PointPillars reported by Lang et al. \cite{lang_pointpillars_2018}}
\begin{center}
\begin{tabular}{ccccccc}
\hline
 & \multicolumn{2}{c}{easy} &  \multicolumn{2}{c}{moderate} &  \multicolumn{2}{c}{difficult}\\
 \hline
 \textbf{car} & mAP & recall & mAP & recall & mAP & recall \\
 \hline
 PointPillars($\dagger$) & 78.17 & 90.0 & 68.71 & 80.0 & 65.91 & 75.0 \\
  + ours & 83.95 & 92.5 & 75.79 & 82.5 & 67.88 & 77.5 \\
 PointPillars($\ddagger$) & 79.05 & 90.0 & 74.99 & 85.0 & 68.30 & 80.0 \\

\hline 
\hline
 \textbf{pedestrian} & mAP & recall & mAP & recall & mAP & recall \\
 \hline
 PointPillars($\dagger$) & 45.67 & 62.5 & 38.65 & 52.5 & 36.16 & 50.0 \\
  + ours & 50.38 & 67.5 & 43.90 & 60.0 & 40.91 & 57.5 \\
 PointPillars($\ddagger$) & 52.08 & 67.5 & 43.53 & 57.5 & 41.49 & 55.0 \\
 
\hline
\hline
 \textbf{cyclist} & mAP & recall & mAP & recall & mAP & recall \\
 \hline
 PointPillars($\dagger$) & 70.92 & 90.0 & 55.57 & 75.0 & 49.95 & 65.0 \\
 + ours & 70.52 & 90.0 & 56.94 & 75.0 & 51.70 & 67.5 \\
 PointPillars($\ddagger$) & 75.78 & 92.5 & 59.07 & 77.5 & 52.92 & 67.5 \\
 
\hline
\hline
\end{tabular}
\end{center}
\label{tab:exp4_test}
\end{table}

\cref{tab:exp4_test} reports evaluation result on KITTI \textit{test} set before/after epBRM is applied. We also reports original PointPillars result($\ddagger$) reported by Lang et al.\cite{lang_pointpillars_2018} for reference. Note that epBRM improves the recall of detection in most cases.


\section{Discussion and future research direction}


In this work, we mainly focused on a box regression task, one of core components in detection pipeline that aims to predict precise 3D bounding box. By adopting spatial transformation mechanism into box regression module, we could build a small network which improves the detection performance within a short time of training.

We summarize the detection failure cases from our experiments. First, epBRM requires that localization modules provide reliable information about the location of objects. If the localization module fails to localize objects precisely, then no further attempt to refine 3D bounding boxes will be made by epBRM. 

Second, given the close proximity between ground truth objects and the predicted locations from localization module, epBRM results in false-positive predictions if the point cloud is sparse. We think that other source of information, such as additional sensor data, can be helpful to overcome such challenge.

The core idea behind the proposed approach is to gradually simplify the box regression task by applying spatial transformation mechanism on raw 3D LiDAR point clouds input. We think that this methodology is generic to other tasks which use raw 3D point clouds and not limited to the box regression task. We will explore the tracking task, optical flow task and instance segmentation task based on the methodology discussed in this work as our future research directions. 



\bibliographystyle{IEEEtran}
\bibliography{references}

\begin{thebibliography}{10}
\providecommand{\url}[1]{#1}
\csname url@samestyle\endcsname
\providecommand{\newblock}{\relax}
\providecommand{\bibinfo}[2]{#2}
\providecommand{\BIBentrySTDinterwordspacing}{\spaceskip=0pt\relax}
\providecommand{\BIBentryALTinterwordstretchfactor}{4}
\providecommand{\BIBentryALTinterwordspacing}{\spaceskip=\fontdimen2\font plus
\BIBentryALTinterwordstretchfactor\fontdimen3\font minus
  \fontdimen4\font\relax}
\providecommand{\BIBforeignlanguage}[2]{{%
\expandafter\ifx\csname l@#1\endcsname\relax
\typeout{** WARNING: IEEEtran.bst: No hyphenation pattern has been}%
\typeout{** loaded for the language `#1'. Using the pattern for}%
\typeout{** the default language instead.}%
\else
\language=\csname l@#1\endcsname
\fi
#2}}
\providecommand{\BIBdecl}{\relax}
\BIBdecl

\bibitem{geiger_vision_2013}
A.~Geiger, P.~Lenz, C.~Stiller, and R.~Urtasun,
  ``\BIBforeignlanguage{en}{Vision meets robotics: {{The KITTI}} dataset},''
  \emph{\BIBforeignlanguage{en}{The International Journal of Robotics
  Research}}, vol.~32, no.~11, pp. 1231--1237, Sep. 2013.

\bibitem{ku_joint_2017}
J.~Ku, M.~Mozifian, J.~Lee, A.~Harakeh, and S.~Waslander, ``Joint {{3D Proposal
  Generation}} and {{Object Detection}} from {{View Aggregation}},''
  \emph{arXiv:1712.02294 [cs]}, Dec. 2017.

\bibitem{Tsung_focal_2017}
\BIBentryALTinterwordspacing
T.~Lin, P.~Goyal, R.~B. Girshick, K.~He, and P.~Doll{\'{a}}r, ``Focal loss for
  dense object detection,'' \emph{CoRR}, vol. abs/1708.02002, 2017. [Online].
  Available: \url{http://arxiv.org/abs/1708.02002}
\BIBentrySTDinterwordspacing

\bibitem{yang_pixor_2018}
B.~Yang, W.~Luo, and R.~Urtasun, ``\BIBforeignlanguage{en}{{{PIXOR}}:
  {{Real}}-{{Time 3D Object Detection From Point Clouds}}},'' in
  \emph{\BIBforeignlanguage{en}{{Proceedings of the IEEE conference on Computer
  Vision and Pattern Recognition}}}, 2018, p.~9.

\bibitem{lang_pointpillars_2018}
A.~H. Lang, S.~Vora, H.~Caesar, L.~Zhou, J.~Yang, and O.~Beijbom,
  ``{{PointPillars}}: {{Fast Encoders}} for {{Object Detection}} from {{Point
  Clouds}},'' \emph{arXiv:1812.05784 [cs, stat]}, Dec. 2018.

\bibitem{li2016vehicle}
B.~Li, T.~Zhang, and T.~Xia, ``Vehicle detection from 3d lidar using fully
  convolutional network,'' \emph{arXiv preprint arXiv:1608.07916}, 2016.

\bibitem{qi_frustum_2018}
C.~R. Qi, W.~Liu, C.~Wu, H.~Su, and L.~J. Guibas, ``Frustum {{PointNets}} for
  {{3D Object Detection}} from {{RGB}}-{{D Data}},'' in \emph{{Proceedings of
  the IEEE conference on Computer Vision and Pattern Recognition}}, 2018.

\bibitem{luo_fast_2018}
W.~Luo, B.~Yang, and R.~Urtasun, ``\BIBforeignlanguage{en}{Fast and
  {{Furious}}: {{Real Time End}}-to-{{End 3D Detection}}, {{Tracking}} and
  {{Motion Forecasting With}} a {{Single Convolutional Net}}},'' in
  \emph{\BIBforeignlanguage{en}{Proceedings of the IEEE conference on Computer
  Vision and Pattern Recognition}}, 2018.

\bibitem{simon_complex-yolo_2018}
M.~Simon, S.~Milz, K.~Amende, and H.-M. Gross, ``Complex-{{YOLO}}:
  {{Real}}-time {{3D Object Detection}} on {{Point Clouds}},'' in
  \emph{{European Conference on Computer Vision}}, Mar. 2018.

\bibitem{cvpr17chen}
X.~Chen, H.~Ma, J.~Wan, B.~Li, and T.~Xia, ``Multi-view 3d object detection
  network for autonomous driving,'' in \emph{Proceedings of the IEEE conference
  on Computer Vision and Pattern Recognition}, 2017.

\bibitem{liang_deep_2018}
M.~Liang, S.~Wang, B.~Yang, and R.~Urtasun, ``\BIBforeignlanguage{en}{Deep
  {{Continuous Fusion}} for {{Multi}}-{{Sensor 3D Object Detection}}},'' in
  \emph{\BIBforeignlanguage{en}{{European Conference on Computer Vision}}},
  2018, p.~16.

\bibitem{enzweiler2011multilevel}
M.~Enzweiler and D.~M. Gavrila, ``A multilevel mixture-of-experts framework for
  pedestrian classification,'' \emph{IEEE Transactions on Image Processing},
  vol.~20, no.~10, pp. 2967--2979, 2011.

\bibitem{gonzalez2017board}
A.~Gonz{\'a}lez, D.~V{\'a}zquez, A.~M. L{\'o}pez, and J.~Amores, ``On-board
  object detection: Multicue, multimodal, and multiview random forest of local
  experts,'' \emph{IEEE transactions on cybernetics}, vol.~47, no.~11, pp.
  3980--3990, 2017.

\bibitem{zining2018fusing}
Z.~{Wang}, W.~{Zhan}, and M.~{Tomizuka}, ``Fusing bird’s eye view lidar point
  cloud and front view camera image for 3d object detection,'' in \emph{2018
  IEEE Intelligent Vehicles Symposium (IV)}, 2018, pp. 1--6.

\bibitem{zhou2017voxelnet}
Y.~Zhou and O.~Tuzel, ``{VoxelNet}: {End}-to-{End} {Learning} for {Point}
  {Cloud} {Based} 3d {Object} {Detection},'' Nov. 2017, arXiv: 1711.06396.

\bibitem{shi2018pointrcnn}
S.~Shi, X.~Wang, and H.~Li, ``Pointrcnn: 3d object proposal generation and
  detection from point cloud,'' \emph{arXiv preprint arXiv:1812.04244}, 2018.

\bibitem{li20173d}
B.~Li, ``3d fully convolutional network for vehicle detection in point cloud,''
  in \emph{2017 IEEE/RSJ International Conference on Intelligent Robots and
  Systems (IROS)}.\hskip 1em plus 0.5em minus 0.4em\relax IEEE, 2017, pp.
  1513--1518.

\bibitem{graham2014spatially}
B.~Graham, ``Spatially-sparse convolutional neural networks,'' \emph{arXiv
  preprint arXiv:1409.6070}, 2014.

\bibitem{vora2019pointpainting}
S.~Vora, A.~H. Lang, B.~Helou, and O.~Beijbom, ``Pointpainting: Sequential
  fusion for 3d object detection,'' \emph{arXiv preprint arXiv:1911.10150},
  2019.

\bibitem{yoo20203d}
J.~H. Yoo, Y.~Kim, J.~S. Kim, and J.~W. Choi, ``3d-cvf: Generating joint camera
  and lidar features using cross-view spatial feature fusion for 3d object
  detection,'' \emph{arXiv preprint arXiv:2004.12636}, 2020.

\bibitem{xiaodeep}
S.~S.~J. Xiao, ``Deep sliding shapes for amodal 3d object detection in rgb-d
  images.''

\bibitem{qi_pointnet_2017}
C.~R. Qi, H.~Su, K.~Mo, and L.~J. Guibas, ``{{PointNet}}: {{Deep Learning}} on
  {{Point Sets}} for {{3D Classification}} and {{Segmentation}},'' in
  \emph{{Proceedings of the IEEE conference on Computer Vision and Pattern
  Recognition}}, 2017.

\bibitem{qi_pointnet++_2017}
C.~R. Qi, L.~Yi, H.~Su, and L.~J. Guibas, ``{{PointNet}}++: {{Deep Hierarchical
  Feature Learning}} on {{Point Sets}} in a {{Metric Space}},'' in
  \emph{{Neural Information Processing Systems}}, 2017.

\bibitem{shin2019roarnet}
K.~Shin, Y.~P. Kwon, and M.~Tomizuka, ``Roarnet: A robust 3d object detection
  based on region approximation refinement,'' in \emph{2019 IEEE Intelligent
  Vehicles Symposium (IV)}.\hskip 1em plus 0.5em minus 0.4em\relax IEEE, 2019,
  pp. 2510--2515.

\bibitem{jaderberg2015spatial}
M.~Jaderberg, K.~Simonyan, A.~Zisserman \emph{et~al.}, ``Spatial transformer
  networks,'' in \emph{Advances in neural information processing systems},
  2015, pp. 2017--2025.

\bibitem{feng2019deep}
D.~Feng, C.~Haase-Schuetz, L.~Rosenbaum, H.~Hertlein, F.~Duffhauss, C.~Glaeser,
  W.~Wiesbeck, and K.~Dietmayer, ``Deep multi-modal object detection and
  semantic segmentation for autonomous driving: Datasets, methods, and
  challenges,'' \emph{arXiv preprint arXiv:1902.07830}, 2019.

\end{thebibliography}

\end{document}